\documentclass[wcp]{jmlr}

\usepackage{longtable}

\usepackage{booktabs}

\makeatletter
\let\Ginclude@graphics\@org@Ginclude@graphics 
\makeatother

\jmlrvolume{157}
\jmlryear{2021}
\jmlrworkshop{ACML 2021}

\title[S2TNet]{S2TNet: Spatio-Temporal Transformer Networks for Trajectory Prediction in Autonomous Driving}

 \author{\Name{Weihuang Chen} \Email{chenweihuang@stu.xjtu.edu.cn}\\
   \Name{Fangfang Wang} \Email{ffwang123@mail.xjtu.edu.cn}\\
   \Name{Hongbin Sun}(corresponding author) \Email{hsun@mail.xjtu.edu.cn}\\
   \addr Xi'an Jiaotong University, Xi'an, China}

\editors{Vineeth N Balasubramanian and Ivor Tsang}

\begin{document}

\maketitle

\begin{abstract}
To safely and rationally participate in dense and heterogeneous traffic, autonomous vehicles require to sufficiently analyze the motion patterns of surrounding traffic-agents and accurately predict their future trajectories. This is challenging because the trajectories of traffic-agents are not only influenced by the traffic-agents themselves but also by spatial interaction with each other. Previous methods usually rely on the sequential step-by-step processing of Long Short-Term Memory networks (LSTMs) and merely extract the interactions between spatial neighbors for single type traffic-agents. We propose the Spatio-Temporal Transformer Networks (S2TNet), which models the spatio-temporal interactions by spatio-temporal Transformer and deals with the temporel sequences by temporal Transformer. We input additional category, shape and heading information into our networks to handle the heterogeneity of traffic-agents. The proposed methods outperforms state-of-the-art methods on ApolloScape Trajectory dataset by more than 7\% on both the weighted sum of Average and Final Displacement Error. Our code is available at \href{https://github.com/chenghuang66/s2tnet}{https://github.com/chenghuang66/s2tnet}.
\end{abstract}


\section{Introduction}
Autonomous driving is an innovative and advanced research field that can reduce the number of road fatalities, increase traffic efficiency, decrease environmental pollution and give mobility to handicapped members of our society (\cite{milakis2017policy}). In order to achieve desired goals and avoid collisions of other agents, autonomous vehicles need to have the ability to perceive the environment and make intelligent decisions. As a part of perception, trajectory prediction can well reflect the future behaviors of surrounding agents and build a bridge between perception and decision-making. However, complex temporal prediction is inevitably accompanied by spatial agent-agent interactions at the same time, especially in the dense and highly dynamic traffic composed of heterogeneous \textit{traffic-agents}, including pedestrians, cyclists, human drivers. The heterogeneity means that these traffic-agents have diverse shapes, sizes, dynamics and behaviors. Moreover, a variety of potentially reasonable spatial interactions between traffic-agents may occur, e.g. human drivers may overtake another vehicle or slow down to follow other vehicles (\cite{lefevre2014survey}). Consequently, trajectory prediction is a challenging task that plays an important role in autonomous driving.

Classical methods treat traffic-agents as individual entities without any spatial interactions and abstract their motion as kinematic and dynamic models (\cite{brannstrom2010model}), Gaussian Processes (\cite{rasmussen2003gaussian}) and etc., making it difficult to comprehend complex scenarios or accomplish long-term predictions. With the success in deep neural networks, recent trajectory prediction methods mainly focus on using these networks to extract features on spatial and temporal dimensions (\cite{alahi2016social}; \cite{huang2019stgat}; \cite{ivanovic2019trajectron}; \cite{mohamed2020social};). Long Short-Term Memory networks (LSTMs) are widely used for modeling temporal features. The LSTMs are based on consecutively processing sequences and storing the latent states to represent knowledge about the motion of traffic-agents (\cite{giuliari2021transformer}). However, LSTM-based methods remember the history with a single vector with limited memory and regularly have difficulty in handling complex temporal dependencies (\cite{vaswani2017attention}). After that, pooling mechanism (\cite{deo2018convolutional}), attention mechanisms (\cite{ivanovic2019trajectron}) and graph convolution mechanisms (\cite{DBLP:journals/corr/abs-1907-07792}; \cite{DBLP:journals/corr/abs-2005-08514}) are used to model the spatial interactions. The limitation of these methods is that they only model the interactions of spatially proximal traffic-agents and ignore the influence by traffic-agents beyond the given spatial limits. This assumption may work well when the speed of traffic-agents is low, but lose efficacy with speed increasing. Besides, the majority of trajectory prediction algorithms are developed for homogeneous traffic-agents in a single scene, which corresponding to human pedestrians in crowds (\cite{alahi2016social}) or moving vehicles on a highway (\cite{deo2018convolutional}. These methods may have great limitation on dealing with dense urban environments where heterogeneous traffic-agents coexist and interact with each other.

In this paper, we address all these limitations by employing Spatio-Temporal Transformer Networks (S2TNet) for heterogeneous traffic-agents trajectory prediction. S2TNet is proposed based on the vanilla Transformer architecture, which discards the sequential nature of data and models features with only the effective self-attention mechanism. For the spatial dimension, we propose spatial self-attention mechanism to capture the interactions between all traffic-agents in the road network, not limited to the interactions between spatial neighbors. For the temporal dimension, temporal convolution network (TCN) is adopted to extract temporal dependencies of consecutive frame and combined with spatial self-attention to form the spatio-temporal Transformer where a set of new spatio-temporal features are obtained. Based on temporal self-attention mechanism, temporal Transformer could refine the temporal features for each traffic-agent independently and produce the future trajectories auto-regressively. In addition to history trajectories, we input additional shape, heading, category features into our networks to handle the heterogeneity of traffic-agents. Main contributions of this paper are summarized as follows:

\begin{itemize}
\item We put forward an innovative approach for heterogeneous traffic-agents trajectory prediction, employing Transformer-based networks to accurately extract interaction information both on the spatial and temporal dimensions. 
\item Spatio-temporal Transformer is designed creatively to merge spatial and temporal information from the history features of traffic-agents. After that, temporal Transformer is utilized to enhance capturing temporal dependencies and output future trajectories with specified length.
\item S2TNet outperforms prior methods on ApolloScape Trajectory dataset by $7.2\%$ on the weighted sum of Average Displacement Error (WSFDE) and $7.7\%$ on the weighted sum of Final Displacement Error (WSFDE)
\end{itemize}

\section{Background}
\subsection{Problem Formulation}
Trajectory prediction aims to accurately predict the future long-term trajectories of traffic-agents, given their history trajectories and other information such as shapes and categories. 

The input of S2TNet is
\begin{equation}
    \textbf{X} = [\textbf{x}^{1},\cdots,\textbf{x}^{t_{obs}}]
\end{equation} where, 
\begin{equation}
    \textbf{x}^i = \{(x^i_0,y^i_0,l^i_0, w^i_0, \theta^i_0, \tau^i_0,\cdots,x^i_n,y^i_n,l^i_n, w^i_n, \theta^i_n, \tau^i_n) \,| \, i\in{(1:t_{obs})}\}
\end{equation}
are the history feature vectors (including global coordinates \textit{x} and \textit{y}, lengths $l$, widths $w$, headings $\theta$ and categories $\tau$) of n traffic-agents being predicted in a road network. The subscript n in (2) refers to all agents in general and varies with different scenes. We currently take into account five types of traffic-agents $c\in{(1,2,3,4,5)}$, representing small vehicles, big vehicles, pedestrian, cyclist and others sequentially. We hold that additional features if available to each traffic-agent could handle the heterogeneity of traffic-agents and improve trajectory accuracy. 

The output of S2TNet is
\begin{equation}
    \textbf{Y} = [\textbf{y}^{t_{obs}},\cdots,\textbf{y}^{t_{fut}}]
\end{equation} where,
\begin{equation}
    \textbf{y}^i = \{(x^i_0,y^i_0,\cdots,x^i_n,y^i_n) \,| \, i\in{(t_{obs+1}:t_{fut})}\}
\end{equation}
are the future feature vectors including global coordinates x and y. It is noted that S2TNet outputs future positions of all observed traffic-agents simultaneously other than merely predicting the location of one specific traffic-agent.

With the objective to hierarchically represent the trajectory sequences, we construct a spatio-temporal graph $G=(V,E)$ on a trajectory sequence with \textit{N} traffic-agents and \textit{T} frames featuring both intra-frame and inter-frame connection. In this graph, the node set $V=\{\textbf{x}^t_i\,| \,t\in(1,T),i\in(1,N)\}$ includes all the feature vectors of traffic-agents, and $E$ represents the set of edges connected between nodes. We utilize node and traffic-agent equally in the following description. The edge set $E$ consists of two subsets. The fist subsets depicts the virtual spatial connection between traffic-agents in the same frame, denotes as  $E_S=\{(\textbf{x}^t_i,\textbf{x}^t_j)\,| \,i,j\in(1,N),t\in(1,T)\}$. The second subset contains the temporal edges which connects the same traffic-agent in consecutive frames as $E_T=\{(\textbf{x}^t_i,\textbf{x}_i{^{t_1}})\,| \,i\in(1,N),t,t{_1}\in(1,T)\}$.

\subsection{Trajectory prediction networks overview}
\noindent \textbf{Trajectory prediction using RNNs}\quad Recurrent Neural Networks (RNNs) and their variant structures such as LSTMs and Gated Recurrent Units (GRU) have made great progress in trajectory prediction. As one of the earliest RNNs using in trajectory prediction, Social-LSTM (\cite{alahi2016social}) addresses interaction between neighborhood by defining a spatial grid based pooling scheme to aggregate the recurrent outputs of all the agents around the agent being predicted. However, this hand-crafted solution is inefficient and fails to capture global context since cells in one grid are treated equally. CS-LSTM (\cite{DBLP:journals/corr/abs-1803-10892}) combines a novel pooling mechanism with generative adversarial networks to tackle intrinsically multimodality of pedestrian trajectory. TrafficPredict (\cite{ma2019trafficpredict}) utilizes LSTM to refine the similarities of motion pattern of instances into category features for heterogeneous agent prediction. SR-LSTM (\cite{zhang2019sr}) introduces a message passing and selecting mechanism to capture the crucial current intention of the neighbors.

\noindent \textbf{Trajectory prediction using hybrid networks}\quad Recently, the approaches of trajectory prediction have been extended to hybrid networks by combining RNNs with Convolution Networks (CNNs), Generative Adversarial Network (GAN) or Graph Neural Networks (GNNs). Traphic (\cite{chandra2019traphic}) introduces CNNs into pooling mechanism for maneuver based trajectory prediction. Sophie (\cite{sadeghian2019sophie}) concatenates the outputs of social and physical attention with the scene features extracted by a CNNs and takes advantage of GAN to generate more realistic samples for the path prediction of multiple interacting agents. Social-BiGAT (\cite{kosaraju2019social}) captures the social interactions information between pedestrians on the basis of graph attention networks. GRIP (\cite{DBLP:journals/corr/abs-1907-07792}) directly models traffic-agents history trajectories as a spatio-temporal graph and forecasts future trajectories based on Spatio-Temporal Graph Convolutional Networks (ST-GCNs).

\noindent \textbf{Trajectory prediction using Transformers}\quad On account of Transformer's (\cite{vaswani2017attention}) unique attention mechanism and superior performance in NLP, there is an emerging interest in applying Transformer architectures to prediction tasks. Without considering any complicated interaction information, (\cite{giuliari2021transformer}) utilizes vanilla Transformer for pedestrian trajectory forecasting and achieves plausible results. STAR (\cite{DBLP:journals/corr/abs-2005-08514}) interleaves a variant of the graph convolution, named TGConv, and the original Temporal Transformer for spatio-temporal interactions modeling. Inspired by the parallel version of transformer used in (\cite{carion2020end}), mmTransformer (\cite{liu2021multimodal}) uses stacked transformers to aggregate multiple information and achieves multimodal prediction.

\subsection{Self-attention in Transformer}
The core of Transformer networks is their unique self-attention mechanism which is used in parallel, while LSTMs only serially combine current word with the embedding of previous words which have been processed. The first step in calculating the self-attention of Transformer is to learn three vectors separately, i.e. query $q_i\in{\mathbb{R}^{d_q}}$, key $k_i\in{\mathbb{R}^{d_k}}$ and value $v_i\in{\mathbb{R}^{d_v}}$ through trainable linear mapping from each embedding $w_i\in{\mathbb{R}^e},i\in(1,n)$, where n is the number of words being considered. After that, a score is calculated by the dot product of a query and key, $\alpha_{ij}=q_i\cdot k_j^T, {\forall}i,j\in(1,n)$, where superscript $T$ is the transpose of vector. Through the softmax function, all scores belonging to the same node are normalized. Finally, the $i$th self-attention is obtained by multiplying each $v_j$ by normalized scores ${\alpha}_{ij}$ and summing the weighted results.

In practice, the attention function is computed on a set of words simultaneously. Three vectors $(q, k, v)$ of all words are individually packed together into three matrices $(Q, K, V)$. The output of this process, named as \textit{scaled dot-product attention}, can be written as:
\begin{equation}
Attention(\textbf{Q,K,V})=softmax(\frac{\textbf{QK}^T}{\sqrt{d_k}})\textbf{V}
\end{equation}
where $d_k$ is the dimension of each query. The division by $\sqrt{d_k}$ is used to increase gradients stability. 

By adding \textit{multi-head attention mechanism}, we can further improve the performance of self-attention. It gives multiple representation sub-spaces for self-attention and enables the model to jointly deal with information from varied sub-spaces at separate positions.

\begin{figure}[pt]
\vspace{-0.8cm}
\begin{center}
\includegraphics[width=1.0\textwidth]{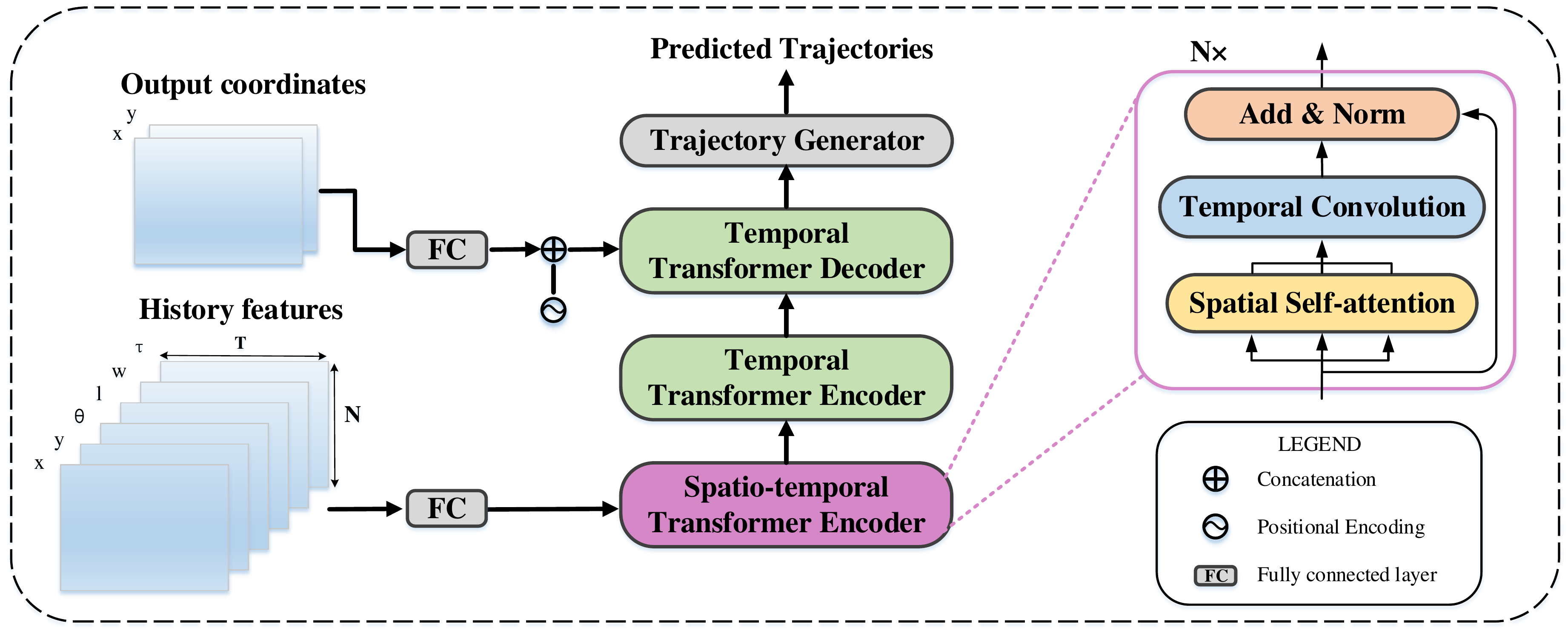}
\caption{Overview of S2TNet. S2TNet leverages the encoders representation of history features, i.e. x and y coordinates, length $l$, width $w$, heading $\theta$ and category $\tau$, of all N traffic-agents in T frames, and the decoder to obtain the refined output spatio-temporal features, and further generates future trajectories by passing them to trajectory generator. The two encoders and one decoder contains a stack of $N = 6$ identical layers respectively. The detailed Temporal Transformer can be found in appendix~\ref{apd:first}.}\label{fig:architecture}
\end{center}
\vspace{-0.3cm}
\end{figure}

\section{Proposed S2TNet Model}
\subsection{Overview of S2TNet Model}
As illustrated in Fig.~\ref{fig:architecture}, the whole model can be viewed as an encoder-decoder architecture in which spatio-temporal Transformer encoder, temporal Transformer encoder and temporal Transformer decoder are aggregated hierarchically. For the sake of acquiring abundant motion information, the history feature vectors of each traffic-agent are embedded onto a higher dimensional space by means of a fully connected layer. Then, the spatial interactions in intra-frame are captured by spatial self-attention and the temporal features of inter-frame are obtained by TCN. Our model emphasizes the coupled spatio-temporal modeling by interleaving the spatial self-attention and TCN in a single spatio-temporal Transformer layer. In order to further capture the temporal dependencies on all history frames, we perform post-processing of the input embeddings with the second temporal Transformer encoder. Temporal Transformer decoder refines the output embeddings based on the spatio-temporal features provided by encoders and the previously predicted output embeddings produced by previously output coordinates. Finally, the trajectory generator outputs all the traffic-agents future trajectories $\textbf{Y}^{(t_{obs}+1,t_{fut})}$ simultaneously by decoding the output embeddings.

\begin{figure}[tp]
\vspace{-0.8cm}
\begin{center}
\includegraphics[width=0.8\textwidth]{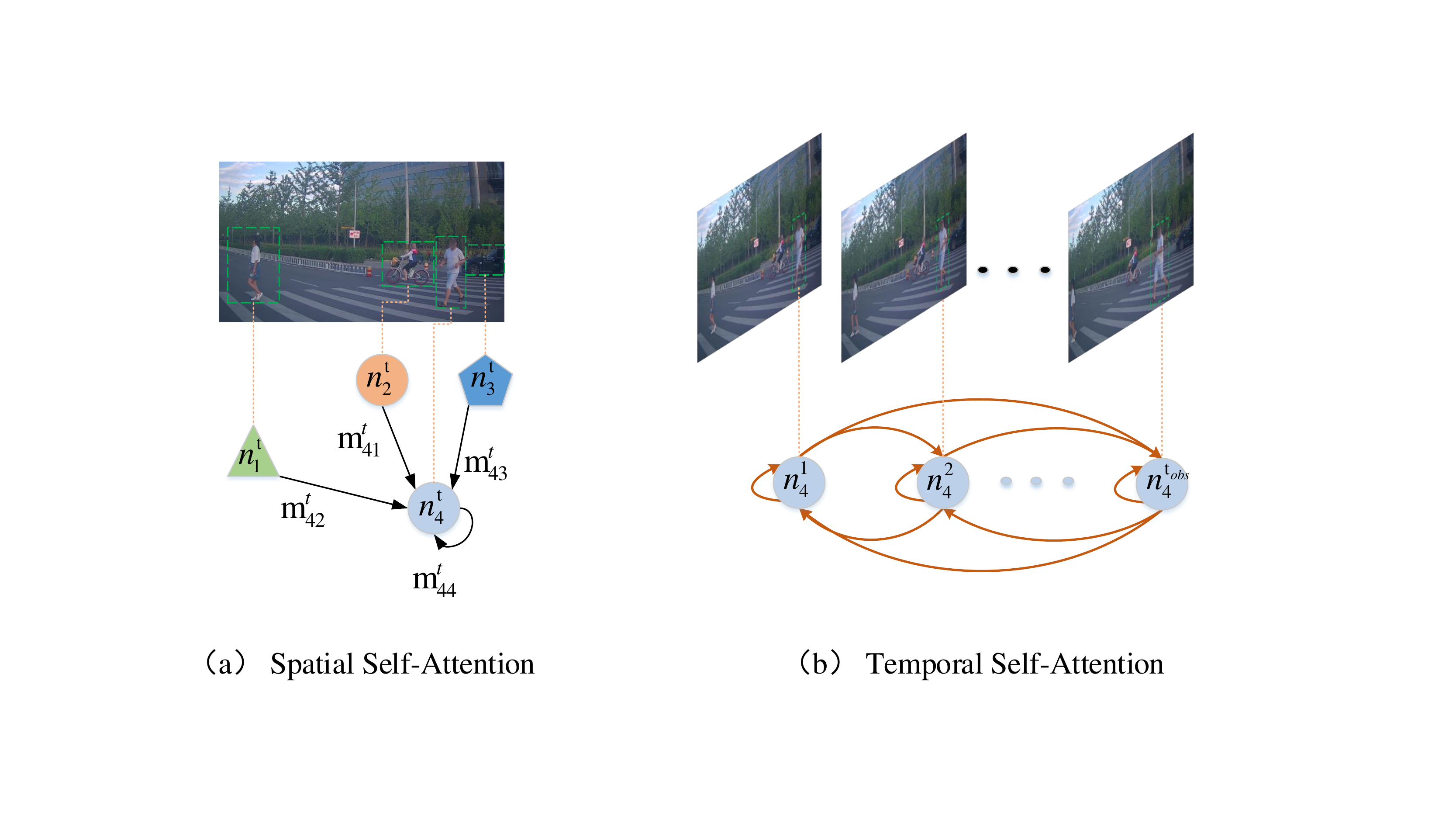}
\caption{Spatial and Temporal Self-Attention. (a) The spatial interactions of node 4 in frame t is modeled. $n^t_i\,(i=1,\, 2,\, 3,\, 4)$ is the embeddings of node \textit{i}.  $m^t_{4j}\,(j=1,\,2,\,3,\,4)$ is the message passing from node j to 4. (b) The temporal correlations between inter-frame are computed in temporal Transformer where the nodes are independent of each other.}\label{fig:graph}
\end{center}
\vspace{-0.3cm}
\end{figure}

\subsection{Spatio-temporal Transformer}
In order to handle the spatial interactions coupled with temporal continuity, we creatively design a spatio-temporal Transformer encoder that captures spatial information through a spatial self-attention sub-layer and extracts dependencies along the temporal dimension through a TCN sub-layer. We interleave two sub-layers to merge the spatio-temporal features.

\noindent \textbf{Spatial Self-attention sub-layer}\quad From a different perspective of Transformer, the spatial attention could be regard as spatial-edge in the spatio-temporal graph. We adopt message passing mechanism on the spatial-edge to preform the suitable processing. For each node $i$ in the scene at time $t$, query $q^t_i\in{\mathbb{R}^{d_q}}$, key $k^t_i\in{\mathbb{R}^{d_k}}$ and value $v^t_i\in{\mathbb{R}^{d_v}}$ is computed by the linear projection from input embeddings $h^t_i\in{\mathbb{R}^{C}}$: 
\begin{equation}
    q^t_i=W_q\cdot{h^t_i},     k^t_i=W_k\cdot{h^t_i},     v^t_i=W_v\cdot{h^t_i}
\end{equation}
where $W_q\in{\mathbb{R}^{C\times{d_q}}},W_k\in{\mathbb{R}^{C\times{d_k}}},W_v\in{\mathbb{R}^{C\times{d_v}}}$. Attention score between node $i$ and $j\in{(1,V)}$ is then obtained by applying scaled dot-product between $q^t_i$ and $k^t_j$, representing the spatial-edge massage $m^t_{ij}$ send from $j$ to $i$, as depicted in Fig.~\ref{fig:graph}(a). 
\begin{equation}
    m^t_{ij} = q^t_i\cdot {k^t_j}^T
\end{equation}
The messages sent from all $j$ to $i$ is normalized over the weights of spatial-edges and summed to get a single attention head of node $i$, as in the following:
\begin{equation}
    head^t_i = \sum_{j}softmax(\frac{m^t_{ij}}{\sqrt{d_k}})v_j
\end{equation}
By repeating this embedding extraction process $h$ times, multi-head attention are concatenated and projected to output embeddings with an fully connected layer:
\begin{equation}
    MultiHead^t_i = \textbf{W}_o\cdot concat([head^t_{i_0},\cdots,head^t_{i_h}])
\end{equation}

\noindent \textbf{Temporal Convolution sub-layer}\quad After spatial information is obtained, we impose temporal convolution operation on the temporal-edge in the spatio-temporal graph to model temporal dynamics within trajectory sequence. Given the input graph of shape $(T,N,C)$, where T is history frame, N is node number and C is the embeddings, we use a standard 2D convolution with the kernel size ($K\times{1}$) to force on processing along the temporal dimension, as expressed in the following:
\begin{equation}
    output_i=Conv_{K\times{1}}(MultiHead^t_i)
\end{equation}

As a Transformer structure, we regularly imply layer normalization (\cite{ba2016layer}) after skip connection in the end of TCN sub-layer. That is, the output of sub-layer is $LayerNorm(x+Sublayer(x))$. In this way, we have a well-defined operation on the constructed spatio-temporal graph.

\subsection{Temporal Transformer}
Temporal Transformer consists of an encoder and decoder. The capability of temporal Transformer encoder is performed to better study the dynamics of each node independently along the temporal dimension. The temporal Transformer decoder is used to refine the output embeddings by encoder outputs and the previously predicted embeddings.

\noindent \textbf{Encoder} Temporal Transformer encoder layer is composed of temporal self-attention sub-layer and separable convolution sub-layer. 
Each encoder layer has two sub-layers. The first sub-layer, called temporal self-attention, uses a multi-head self-attention mechanism similar to spatial self-attention sub-layer in Spatial Transformer with the difference that correlations along the temporal dimension are computed independently for each node. As shown in Fig.~\ref{fig:graph}(b), the temporal self-attention for node i represented as:
\begin{equation}
    {MultiHead}_i = \textbf{W}_u\cdot{concat([head_{i_0}},\cdots,head_{i_h}])
\end{equation}
\begin{equation}    
    where,{head}_i = softmax(\frac{Q_{i}\cdot{K}^T_{i}}{\sqrt{d_k}})V_i
\end{equation}
Where $Q^i, K^i$ and $V^i$ are query, key and value matrix learned from the embeddings of input node $i$.

Instead of fully connected network used in vanilla Transformer, the second sub-layer is the separable convolution (\cite{DBLP:journals/corr/Chollet16a}) in order to achieve higher accuracy. 

\noindent \textbf{Decoder} To inject the relative position information of previous output trajectories to decoder, we add the positional encodings to output embeddings: 
\begin{equation}
    PE_{(pos,2i)}=sin(pos/10000^{2i/d_{model}})
\end{equation}
\begin{equation}
    PE_{(pos,2i)}=cos(pos/10000^{2i/d_{model}})
\end{equation}
where $pos$ is the position, $i$ is the dimension and $d_{model}$ the total dimensions of the output embeddings.

Compared with temporal self-attention in encoder, decoder employs a masked temporal self-attention sub-layer to ensure that the predictions for time $t$ can only depend on the known outputs at times less than $t$. Besides masked temporal self-attention and separable convolution, a third sub-layer is inserted into decoder layer which performs multi-head attention over the output of the temporal Transformer encoder. 

\subsection{Implementation Details}
The scheme is implemented using PyTorch. The dimensions of embedding features is set to 32. We apply dropout to the output of each sub-layer before the skip connection step and the output of positional encodings in the decoder stacks. The dropout ratio is $P_{drop}=0.1$. 

An L2-loss is adopted
\begin{equation}
    Loss=\sum^T_{t=t_{obs}+1}{|Y^t_{pred}-Y^t_{GT}|^2}
\end{equation}
where $Y^t_{pred}$ and $Y^t_{GT}$ are predicted positions and ground truth positions respectively. We use Adam \cite{kingma2014adam} as the optimizer and impose a learning rate variation strategy as follows:
\begin{equation}
    learning\_rate=d^{-0.5}_{dmodel}\cdot{min(step\_num^{-0.5},step\_num\cdot{warmu\_steps^{-1.5}})}
\end{equation}
where $warmup\_step$ is set to 5000. Random rotation is implemented for data augmentation in the training.

\section{Experiments}
\subsection{Dataset and Evaluation Metrics}
Our model is evaluated on ApolloScape Trajectory dataset (\cite{ma2019trafficpredict}) which is collected by Apollo autonomous vehicles. The ApolloScape Trajectory dataset contains images, point clouds, and manually annotated trajectories. It is gathered under various lighting conditions and traffic densities in Beijing, China. More specifically, it comprises vastly complex traffic flows mixed with vehicles, riders, and pedestrians. The dataset includes 53 minute training sequences and 50 minute testing sequences captured at 2 frames per second. We need to predict six future frames based on six history frames. Due to the testset of ApolloScape Trajectory dataset is not public, we obtain the results of our model and other baselines by uploading to the ApolloScape Trajectory Leaderboard $\footnote{http://apolloscape.auto/leader\_board.htmll}$.

Two metrics are used to evaluate model performance: the Average Displacement Error (ADE) (\cite{5459260}) and the Final Displacement Error (FDE). ADE is the mean Euclidean distance over all predicted positions and ground truth positions during the prediction time, and FDE is the last item of ADE. Obviously, ADE shows the average prediction performance, while the FDE reflects just the prediction accuracy at the end points. Because the trajectories of heterogeneous traffic-agents are
diverse in scales, we use the following weighted sum of ADE (WSADE) and weighted sum of FDE (WSFDE) as metrics:
\begin{equation}
    WSADE=D_v\cdot{ADE_v}+D_p\cdot{ADE_p}+D_b\cdot{ADE_b}
\end{equation}
\begin{equation}
    WSFDE=D_v\cdot{FDE_v}+D_p\cdot{FDE_p}+D_b\cdot{FDE_b}
\end{equation}
where $D_v=0.20$, $D_p=0.58$, and $D_b=0.22$ are relevant with reciprocals of the average velocity of vehicles, pedestrian and cyclist in the dataset.

\subsection{Baselines} \label{baseline}
To evaluate the performance of S2TNet, we compare S2TNet with a wide range of baselines, including:
\begin{itemize}
    \item \textit{Constant Velocity (CV)}: We use the average velocity of history trajectories as the constant velocity during the future to predict trajectories.
    \item \textit{TrafficPredict}: A LSTM-based method using a hierarchical architecture by (\cite{ma2019trafficpredict}).
    \item \textit{StarNet}: (\cite{DBLP:journals/corr/abs-1906-01797}) builds a star topology to consider the collective influence among all pedestrians.
    \item \textit{Social LSTM (S-LSTM)}: (\cite{alahi2016social}) uses LSTM to extract single pedestrian feature and devises a social pooling mechanism to capture neighbor information.
    \item \textit{Social GAN (S-GAN)}: (\cite{DBLP:journals/corr/abs-1803-10892}) predicts socially plausible futures by a conditional GAN. 
    \item \textit{Transformer}: (\cite{giuliari2021transformer}) uses vanilla temporal Transformer to model pedestrian separately without any complex human-human interactions nor scene interaction terms.
    \item \textit{STAR}: (\cite{DBLP:journals/corr/abs-2005-08514}) interleaves spatial and temporal Transformer to capture the social intersection between pedestrians.
    \item \textit{TPNet}: (\cite{Fang_2020_CVPR}) first generates a candidate set of future trajectories, then gets the final predictions by classifying and refining the candidates.
    \item \textit{GRIP++}: (\cite{DBLP:journals/corr/abs-1907-07792}) is the SOTA trajectory predictor which uses a enhanced graph to represent the interactions of close objects, and applies ST-GCNS to extract spatio-temporal features.
\end{itemize}

\subsection{Quantitative Results and Analyses}
We compare S2TNet with the state-of-the-art approaches as mentioned in Section~\ref{baseline}. All methods are compared by the results released on ApolloScape Trajectory Leaderboard. The main results are presented in Table~\ref{tab1}. 

From Table~\ref{tab1} we observe that the performance of S2TNet is superior to the baseline methods of all traffic-agent types by a large margin. More specifically, our method reduces the ADE of vehicles, pedestrians, and cyclists over GRIP++ by 11.28$\%$, 4.31$\%$ and 10.24$\%$ respectively. Meanwhile, our method reduces the FDE of vehicles, pedestrians, and cyclists over GRIP++ by 12.21$\%$, 4.98$\%$ and 5.87$\%$ sequentially. It is notice worthy that the degree of improvement in vehicles and cyclists is better than pedestrians. We believe it is because that the motion pattern of pedestrians are more flexible than vehicles and bikes with non-holonomic constraint. Another remarkable finding is that simple model CV which just makes use of average velocity of history trajectories outperforms many deep learning methods including the SOTA model, STAR. This suggests that homogeneous methods may not handle dense urban scenes efficiently. On the contrary, our approach performs better in heterogeneous and dense urban environments. We will further demonstrate this in Section~\ref{visualize} with visualized results.

\begin{table}[htbp]
\caption{Comparison with baselines models on ApolloScape Trajectory dataset.}
\centering
\begin{center}
\setlength{\tabcolsep}{1.2mm}
\begin{tabular}{c|c|ccc|c|ccc}
\hline
Method          &WSADE              &ADEv&ADEp&ADEb&WSFDE                       &FDEv&FDEp&FDEb    \\ \hline \hline
TrafficPredict & 8.5881          & 7.9467 & 7.1811 & 12.8805 & 24.2262         & 12.7757 & 11.121 & 22.7912 \\ \cline{1-2} \cline{6-6}
S-LSTM         & 1.8922          & 2.9456 & 1.2856 & 2.5337  & 3.4024          & 5.2802  & 2.3240 & 4.5384  \\ \cline{1-2} \cline{6-6}
S-GAN          & 1.5829          & 3.0430 & 0.9836 & 1.8354  & 2.7796          & 5.0913  & 1.7264 & 3.4547  \\ \cline{1-2} \cline{6-6}
STAR           & 1.5400          & 2.5644 & 0.9473 & 2.1714  & 2.8602          & 4.6324  & 1.8029 & 4.0366  \\ \cline{1-2} \cline{6-6}
CV             & 1.4762          & 2.6454 & 0.8547 & 2.0519  & 2.7601          & 4.7944  & 1.6428 & 3.8564  \\ \cline{1-2} \cline{6-6}
StraNet        & 1.3425          & 2.3860 & 0.7854 & 1.8628  & 2.4984          & 4.2857  & 1.5156 & 3.4645  \\ \cline{1-2} \cline{6-6}
Transformer    & 1.2803          & 2.2322 & 0.7398 & 1.8398  & 2.4024          & 4.0317  & 1.4309 & 3.4826  \\ \cline{1-2} \cline{6-6}
TPNet          & 1.2800          & 2.2100 & 0.7400 & 1.8500  & 2.3400          & 3.8600  & 1.4100 & 3.4000  \\ \cline{1-2} \cline{6-6}
GRIP++         & 1.2588          & 2.2400 & 0.7142 & 1.8024  & 2.3631          & 4.0762  & 1.3732 & 3.4155  \\ \hline 

S2TNet   & \textbf{1.1679} & \textbf{1.9874} & \textbf{0.6834} & \textbf{1.7000}  & \textbf{2.1798} & \textbf{3.5783}  & \textbf{1.3048} & \textbf{3.2151}  \\ \hline
\end{tabular}
\label{tab1}
\end{center}
\end{table}

\subsection{Qualitative Results and Analyses} \label{visualize}
\begin{figure}[htp]
\begin{center}
\includegraphics[width=1.0\textwidth]{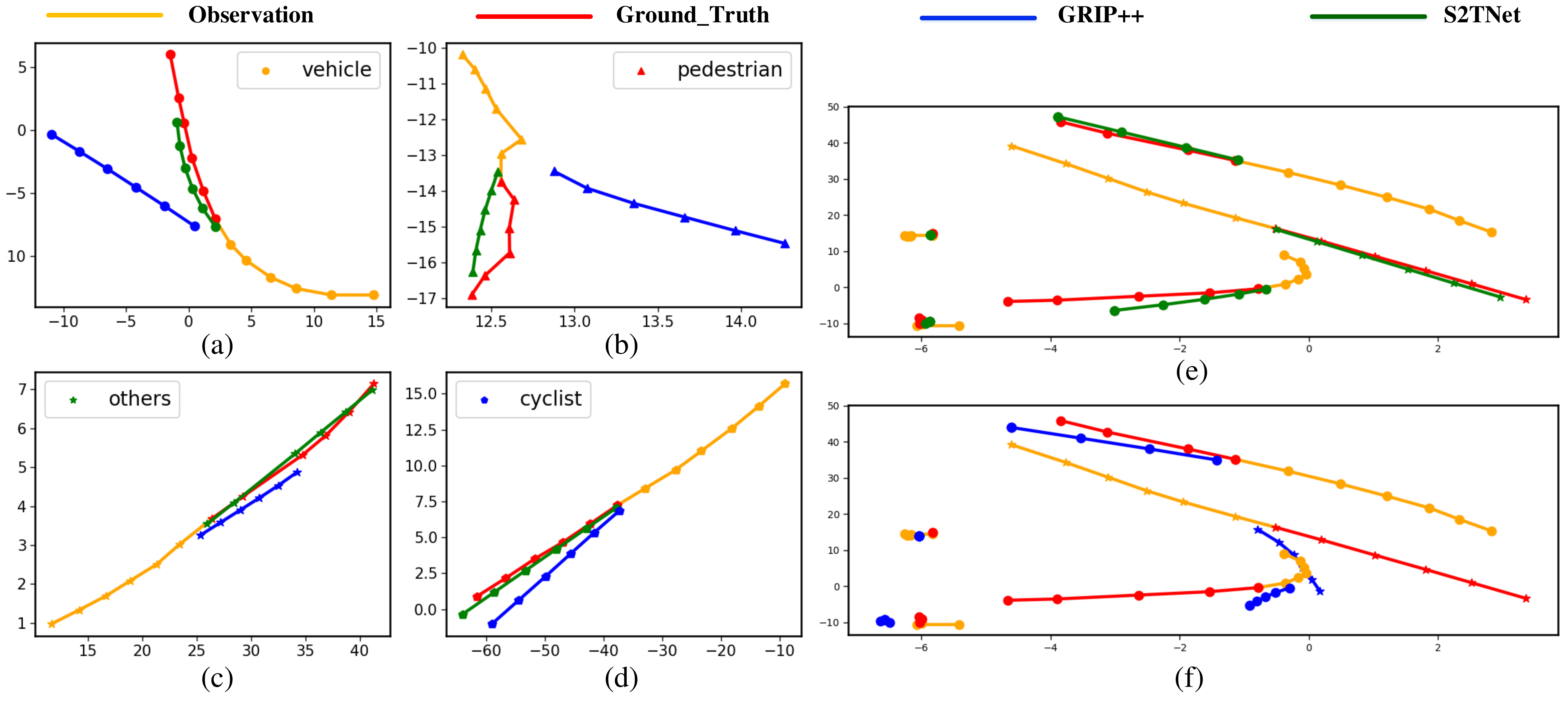}
\caption{Visualized Prediction Results in heterogeneous and dense traffic. S2TNet successfully captures spatio-temporal information and outperforms the SOTA model, GRIP++. (a, b, c, d) Comparison the future trajectories of different types of traffic-agents between two methods. (e, f) The prediction results of GRIP++ and S2TNet in a complete traffic scene.}
\label{fig:experiment}
\end{center}
\vspace{-0.3cm}
\end{figure}

In Fig.~\ref{fig:experiment}, we visualize several prediction results of ApolloScape Trajectory dataset. We separately present the trajectory of single traffic-agent with different type selected from complex scenes and show the complete prediction results of a scene.
\begin{itemize}
    \item \textit{S2TNet has the ability to forecast long horizon trajectories for different categories of traffic-agents.} After observing 6 frames (3s) of history trajectories, S2TNet could accurately predict the trajectories over 3 seconds horizon. Moreover, S2TNet does well in the case of sharp turns for the vehicle, e.g. Fig.~\ref{fig:experiment}(a) and (b). With the increase of prediction length, the prediction results of S2TNet are more realistic and the cumulative error of S2TNet is better than GRIP++, e.g. Fig.~\ref{fig:experiment}(c) and (d).
    \item \textit{S2TNet is able to model spatio-temporal interaction accurately.} In the top right portion of Fig.~\ref{fig:experiment}(e) and (f), a vehicle runs in opposite directions to an unknown traffic-agent. While the predicted trajectories of GRIP++ deviates from ground truth, S2TNet precisely captures the interactive routes.
    \item \textit{S2TNet successfully identify the stationary traffic-agent.} In the lower-left of Fig.~\ref{fig:experiment}(e) and (f), two vehicles decelerate to near standstill. Compared with GRIP++, S2TNet successfully predicts the corresponding stationary trajectories.
\end{itemize}

\subsection{Ablation Studies}
In this section, we conduct extensive ablation studies and focus on the effect of the proposed components. The results are presented in Table~\ref{tab2}.
\begin{itemize}
    \item \textit{The spatio-temporal Transformer could sufficiently extract information both in spatial and temporal dimensions.} In (1), (2) and (3), we remove one or two sub-layers in spatio-temporal Transformer. Compared (1) to (2), the model contains TCN sub-layer outperforms solely temporal Transformer. On the contrast to outperforming in our validation set, (3) which contains the spatial self-attention sub-layer and temporal Transformer is worse than (1) in final test set. We hold that merely stacking attention on the spatial dimension without merging temporal information results in overfitting.

    \item \textit{The temporal Transformer encoder enhance capturing temporal dependencies.} In (4), we remove the temporal Transformer encoder and gain a lower performance compared with (8). This indicates that temporal self-attention mechanism could effectively improve the ability to extract temporal information.
    \item \textit{The separable convolution outperforms full connected feed-forward network in temporal Transformer.} In (5), we replace separable convolution sub-layer in temporal Transformer with full connected feed-forward network and the performance slightly descends.
    \item \textit{More features, higher accuracy.} Instead of feeding all features into S2TNet, we input only history trajectories in (6). We find that rich information helps the network to understand the heterogeneity of traffic-agents.
    \item \textit{The spatial self attention of the whole scene is better than that of the given spatial limits} We use a masked attention mechanism in (7) to ignore the influence out of the given spatial limits (15m) as (\cite{DBLP:journals/corr/abs-1907-07792}) does. We find that the traffic-agents in the whole scene have a great influence on the accuracy of trajectory prediction.

\end{itemize}

\begin{table}[htbp]
\caption{Ablation study. \textbf{SS} denotes spatial self-attention sub-layer in spatio-temporal Transformer. \textbf{TE} denotes temporal Transformer encoder layer. \textbf{SC} denotes separable convolution. \textbf{FC} denotes full connected layer. \textbf{TD} denotes temporal Transformer decoder layer. \textbf{HF} denotes history features. \textbf{A} denotes history features including global coordinates, category, length, width and heading. \textbf{C} denotes only global coordinates. \textbf{LM} denotes spatial limits used in spatial self-attention sub-layer. \textbf{W} denotes the spatial self-attention without spatial limits. \textbf{N} denotes the spatial self-attention of neighbors (15m).}
\centering
\begin{center}
\begin{tabular}{ccccccc|c}
\hline
\multicolumn{7}{c|}{Components}  & Performance   \\ \cline{1-7}
    & SS            & TCN            & TE       & TD    & HF     & LM       & (WSADE/WSFDE) \\ \hline \hline
(1) &$\times$       &$\times$        & SC       & SC    & A      &W   & 1.2300/2.2949 \\
(2) &$\times$       & \checkmark     & SC       & SC    & A      &W       & 1.2189/2.2570              \\
(3) &\checkmark     &$\times$        & SC       & SC    & A      &W       & 1.2500/2.3561              \\
(4) &\checkmark     &\checkmark      &$\times$  & SC    & A      &W       & 1.2674/2.4086 \\
(5) &\checkmark     &\checkmark      & FC       & FC    & A      &W       & 1.1945/2.2613 \\
(6) &\checkmark     &\checkmark      & SC       & SC    & C    &W       & 1.2170/2.3036 \\
(7) &\checkmark     &\checkmark      & SC       & SC    & A     &N       & 1.2686/2.3548 \\
(8) &\checkmark     &\checkmark      & SC       & SC    & A      &W       & \textbf{1.1679/2.1798} \\ \hline
\end{tabular}
\label{tab2}
\end{center}
\end{table}

\section{Conclusion}
In this paper, we propose S2TNet, a Transformer-based framework to predict the trajectories of heterogeneous traffic-agents around autonomous driving cars. Spatio-temporal Transformer is designed to capture spatio-temporal interactions between all traffic-agents, not limited to spatial neighbor. The temporal Transformer is utilized to enhance modeling temporal dependencies and output future trajectories auto-regressively. The experimental results from ApolloScape Trajectory dataset show that the proposed method achieves the state-of-the-art performance and substantially  improves the accuracy of the predicted trajectories. In the future work, we intend to integrate additional map information on S2TNet framework and implement real time prediction on autonomous driving platform by S2TNet.

\acks{This research is supported by National Natural Science Foundation of China (No. 61790563).}

\bibliography{acml21}

\appendix

\section{Temporal Transformer encoder and decoder}\label{apd:first}
\begin{figure}[htp]
\begin{center}
\includegraphics[width=0.6\textwidth]{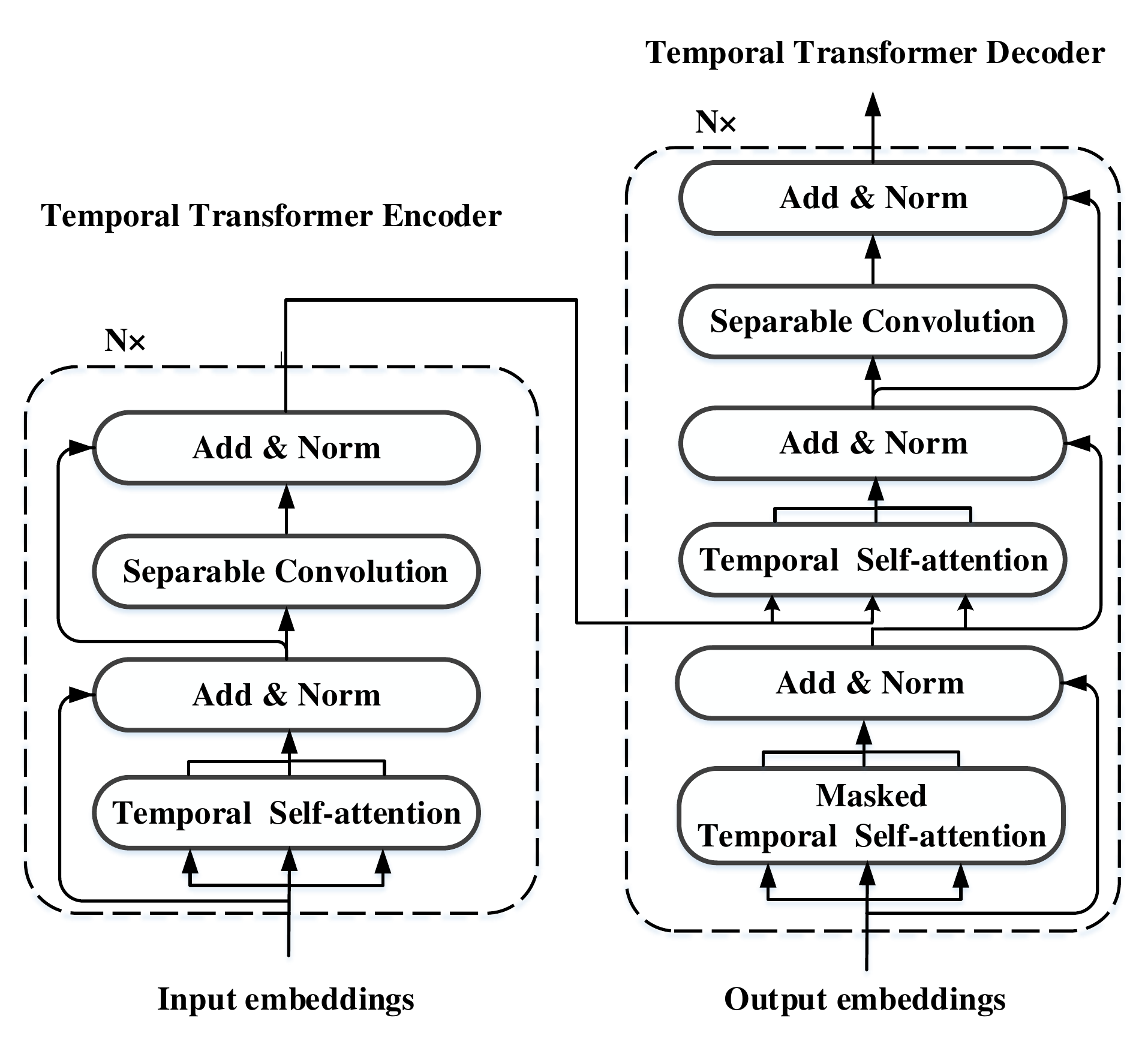}
\caption{Temporal Transformer encoder and decoder}
\label{fig:temp}
\end{center}
\vspace{-0.8cm}
\end{figure}

The detailed temporal Transformer architecture used in S2TNet is visualized in Fig.~\ref{fig:temp}. Input embeddings is passed to the temporal Transformer encoder to enhance capturing the temporal features of observed traffic-agents. Then, the temporal Transformer decoder receives the previously output embeddings and produced the refined output embeddings through masked temporal self-attention, decoder-encoder attention and separable convolution layers.

\end{document}